\documentclass[conference]{ieee/IEEEtran}
\usepackage[utf8]{inputenc}
\usepackage[cmex10]{amsmath}
\usepackage{amsfonts,amsthm}
\usepackage{amssymb,mathrsfs}
\usepackage{graphicx}
\usepackage[caption=false,font=footnotesize]{subfig}
\usepackage{cite}
\usepackage{array,multirow}
\usepackage{hyperref}

\title{An evolutionary approach to the identification of Cellular Automata based on partial observations}  

\author{
\IEEEauthorblockN{Witold Bołt\IEEEauthorrefmark{2}\IEEEauthorrefmark{1}, 
Jan M. Baetens\IEEEauthorrefmark{1} and
Bernard De Baets\IEEEauthorrefmark{1}}
\IEEEauthorblockA{\IEEEauthorrefmark{2}Systems Research Institute, Polish Academy of Sciences, Warsaw, Poland}
\IEEEauthorblockA{\IEEEauthorrefmark{1}KERMIT, Department of Mathematical Modelling, Statistics and Bioinformatics, Ghent University, Ghent, Belgium}
}

\DeclareMathOperator{\fit}{fit}

\DeclareMathOperator{\com}{com}

\DeclareMathOperator{\dist}{dist}

\DeclareMathOperator{\rules}{\mathcal{R}}

\newtheorem{prop}{Proposition}

\newtheorem{fact}[prop]{Fact}

\theoremstyle{remark}
\newtheorem{example}{Example}

\usepackage{mathtools}

\DeclarePairedDelimiter\abs{\lvert}{\rvert}

\begin{document}
\IEEEoverridecommandlockouts
\IEEEpubid{\makebox[\columnwidth]{978-1-4799-7492-4/15/\$31.00~
\copyright2015
IEEE \hfill} \hspace{\columnsep}\makebox[\columnwidth]{ }}
\maketitle

\begin{abstract}
In this paper we consider the identification problem of Cellular Automata (CAs). The problem is defined and solved in the context of partial observations with time gaps of unknown length, {\it i.e.}\ pre-recorded, partial configurations of the system at certain, unknown time steps. A solution method based on a modified variant of a Genetic Algorithm (GA) is proposed and illustrated with brief experimental results.
\end{abstract}

\section{Introduction}
\label{sec:intro}
CAs present an attractive and effective modelling technique for a variety of problems. In order to use CAs in a practical modelling task, one needs to understand the underlying rules, relevant to the given phenomenon, and translate them into a CA local rule. Additionally, the state space, tessellation and neighborhood structure need to be pinned down beforehand. This narrows the application area for CAs, since there are problems for which it is hard to manually design a proper local rule. In some cases only the initial and final states of the system are known ({\it e.g.} \cite{urban-ga,pattern-ga,image-proc}). Such problems motivate the research on automated CA identification. Various methods have been used, including genetic algorithms (GAs) \cite{richards1990extracting,mitchell1996evolving,foo01,Sapin:2003:RCA:1762668.1762709}, genetic programming \cite{conf/automata/BandiniMV08,DBLP:journals/jca/MaedaS07,andre1996discovery}, gene expression programming \cite{DBLP:journals/corr/cs-AI-0102027}, ant intelligence \cite{Liu:2008:BAD:1459980.1459984}, machine learning \cite{DBLP:journals/jca/BullA07}, as well as direct search/construction approaches \cite{adamatzky1994identification,journals/tsmc/YangB00,Yang:2000:NDR:2229236.2229683,Sun11}. 

Existing methods can be divided into two main groups. Firstly, methods for solving specific, global problems. An example of such a problem is majority classification in which one only knows the initial condition and the desired outcome. Secondly, methods that exploit the entire time series of configurations, where it is assumed that all configurations are known. Only limited research efforts have been devoted to problems involving identification based on partial information \cite{richards1990extracting}. 

The main goal of the research presented in this paper is to develop methods capable of automated CA identification in case of partial information. The paper is organized as follows. In Section \ref{sec:prem} we start with introducing basic definitions and presenting some well-known facts on CAs. Section \ref{sec:prob} holds the formal definition of the CA identification problem, while in Section \ref{sec:opt} we reformulate this problem as an optimization task. In Section \ref{sec:ga} the evolutionary algorithm for solving the identification problem is presented. The paper is concluded by Section \ref{sec:res} which presents initial results of computational experiments.   

An introduction to the methods presented in this paper, and a simpler formulation of the discussed algorithm can be found in \cite{wbolt-automata2013}.

\section{Preliminaries}
\label{sec:prem}
\IEEEpubidadjcol
We start by defining a CA. In this paper we will concentrate on 1D, deterministic CAs with a symmetric neighborhood. 

Let $r\in\mathbb{N}$ and $f_A:\{0,1\}^{2\,r+1}\to \{0,1\}$ be any function, then for $N > 0$ we define the $N$--cell global CA rule $A_N\colon\{0,1\}^N\to\{0,1\}^N$ as:
\begin{equation}
A_N(\dotsc,s_i,\ldots) = (\dotsc,f_A(s_{i-r},\dotsc,s_{i+r}),\ldots),
\end{equation}
using periodic boundary conditions, {\it i.e.} for any $i\in\mathbb{Z}$ it holds that $s_{i+N} = s_{i}$. 

The function $f_A$ used in this definition will be referred to as a local rule, and the integer $r$ will be referred to as the radius of the neighborhood. Any local rule can be uniquely defined by a lookup table (LUT) that lists all of the possible arguments together with the corresponding function values. It is assumed that the arguments are listed in a lexicographic order. The general form of such a LUT in the case of radius $r=1$ is shown in Table~\ref{tab:lut-example}.

\begin{table}[ht]
\renewcommand{\arraystretch}{1.3}
\caption{LUT of local rule $R = (l_8,l_7,l_6,l_5,l_4,l_3,l_2,l_1)_2$}
\label{tab:lut-example}
\centering
\begin{tabular}{|>{$}c<{$}|>{$}c<{$}|>{$}c<{$}|>{$}c<{$}|>{$}c<{$}|>{$}c<{$}|>{$}c<{$}|>{$}c<{$}|}
\hline
111 & 110 & 101 & 100 & 011 & 010 & 001 & 000 \\
\hline
l_8 & l_7 & l_6 & l_5 & l_4 & l_3 & l_2 & l_1 \\
\hline
\end{tabular}
\end{table}

The LUT can be used to enumerate local rules, as the coefficients $l_i$ can be treated as digits in the binary representation of an integer $R$, {\it i.e.}\ $R=\sum_{i=1}^8 l_i\,2^{i-1}$. Clearly this extends to higher values of the radius. Due to the fact that the ordering of arguments in the LUT is fixed, only the second row needs to be stored, such that a LUT may be represented as a binary vector. The length of such a vector is $2^{2\,r+1}$. 

With $\{0,1\}^\ast$ we will denote the set of all binary sequences of finite length, {\it i.e.}\ $\{0,1\}^\ast = \bigcup_{N>0}\{0,1\}^N$. The function $A\colon \{0,1\}^\ast\to\{0,1\}^\ast$, satisfying $A(X) = A_N(X)$ if and only if $X\in\{0,1\}^N$, where each of the global rules $A_N$ is defined with the same local rule $f_A$, will be referred to as a generalized global rule of a CA. Such functions will be frequently used throughout this paper, therefore we will simply refer to them as global rules or rules. In this paper a CA will be identified in terms of its global rule, and  by referring to a CA we therefore always refer to its global rule in this generalized sense. 

Note that rule $A$ is uniquely defined by a given local rule $f_A$, but the opposite is not true. For a given rule $A$ we may find different local rules defining it. Fact \ref{fac:samelocal} highlights the relationship between different local rules defining the same~CA.

\begin{fact}
Two local rules $f\colon\{0,1\}^{2\,r+1}\to \{0,1\}$ and $g\colon\{0,1\}^{2\,u+1}\to\{0,1\}$, $u\leq r$, define the same CA if and only if it holds:
\begin{equation}
f(s_1,\dotsc,s_{2\,r+1}) = g(s_{r-u+1},\dotsc,s_{r+u+1}),
\end{equation}
for any $(s_1,\dotsc,s_{2\,r+1})\in\{0,1\}^{2\,r+1}$.\label{fac:samelocal}  
\end{fact}

\begin{example}
Let $g\colon \{0,1\} \to \{0,1\}$ be defined by $g(s) = s$ and $f\colon \{0,1\}^3 \to \{0,1\}$ be defined by $f(s_1, s_2, s_3) = s_2\,s_3 + s_2\,(1-s_3)$. We can see that for any $s_1, s_3\in\{0,1\}$ it holds that $f(s_1, s_2, s_3) = g(s_2) = s_2$, and thus $f$ and $g$ define the same CA rule, which happens to be the identity rule.\qed
\end{example}

For a given neighborhood radius $r$, $\mathcal{A}_r$ denotes the set of all CAs that can be expressed with the use of a local rule with a neighborhood of radius $r$. CAs belonging to $\mathcal{A}_1$ are referred to as Elementary CAs (ECAs), and form the most commonly studied class of 2--state CAs \cite{RevModPhys.55.601}.

Two important properties of the sets $\mathcal{A}_r$ are underlined in Fact \ref{fac:incl}. 

\begin{fact}
For any $r\geq 0$, $\mathcal{A}_r \subset \mathcal{A}_{r+1}$ and $\abs{\mathcal{A}_r} = 2^{2^{2\,r+1}}$.
\label{fac:incl}\end{fact}

Let $A$ be a CA, $X\in \{0,1\}^M$ for some $M$ and $T>0$. The finite sequence of vectors given by:
\[(X, A(X), A^2(X)), \dotsc, A^{T-1}(X)),\]
where $A^t$ denotes the $t$--th application of the rule $A$, will be referred to as the space-time diagram containing $T$ time steps. Each of the elements of a space-time diagram will be referred to as a configuration of the CA, while the first element will be referred to as the initial configuration. If $t=0,1,\dotsc,T-1$ and $m=1,\dotsc,M$, then $A^t(X)[m]$ refers to the state of the $m$--th cell in the $t$--th element of the space-time diagram.  

\begin{example}
We consider an ECA defined by local rule 150. The LUT of ECA 150 is shown in Table \ref{tab:lut-150}.

\begin{table}[ht]
\renewcommand{\arraystretch}{1.3}
\caption{LUT of ECA 150}
\label{tab:lut-150}
\centering
\begin{tabular}{|>{$}c<{$}|>{$}c<{$}|>{$}c<{$}|>{$}c<{$}|>{$}c<{$}|>{$}c<{$}|>{$}c<{$}|>{$}c<{$}|}
\hline
111 & 110 & 101 & 100 & 011 & 010 & 001 & 000 \\
\hline
1 & 0 & 0 & 1 & 0 & 1 & 1 & 0 \\
\hline
\end{tabular}
\end{table}

Figure \ref{fig:spatio1} depicts a space-time diagram of ECA 150, starting from a random initial configuration. Following a common convention, the space-time diagram is visualized as a bitmap in which every row corresponds to a configuration at specific time step. The first row in the image is the initial configuration. State one is drawn as a black pixel, while white pixel corresponds to state zero.\qed
\begin{figure}[!t]
\centering
\includegraphics[width=2.5in]{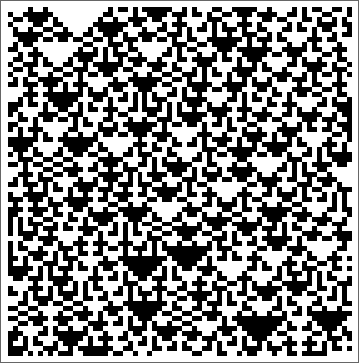}
\caption{Space-time diagram of ECA 150}
\label{fig:spatio1}
\end{figure}
\end{example}

\section{Problem statement}
\label{sec:prob}

In this section we define the identification problem. The formulation presented below is based on the concept of an observation of a space-time diagram, which is assumed to be incomplete, {\it i.e.}\ it contains only partial information on the states of the CA.

Formally, we assume that the states of a system, which is assumed to be an unknown CA, were observed at certain, unknown time steps. Let $I$ be an $N\times M$ array containing symbols belonging to the set $\{0,1,?\}$, where the symbols $0$ and $1$ denote valid states, while $?$ denotes an unknown state belonging to the set $\{0,1\}$. Additionally, let the first row $I[1]\in \{0,1\}^M$. Such an array $I$ will be referred to as an observation. If an observation $I$ does not contain the symbol $?$, we refer to it as spatially complete. The first row $I[1]$ is assumed to represent the initial configuration of a CA, and row $I[n]$ for $n>1$ represents the configuration at time step $\tau_n$. It is assumed that $\tau_n < \tau_{n+1}$.  

Let $I$ be an observation. The number $C(I) = \#\{I[n,m]\neq\ ?\}$ will be referred to as the number of completely observed states. In our case, for any observation $I$ it holds that $C(I)>0$. 

For each observation $I$, we define the set $\com(I)$ that contains all of the spatially complete observations $I'$, satisfying $I'[n,m] = I[n,m]$ for all $n,m$ such that $I[n,m]\neq\,?$.

\begin{example}
\label{ex:i}
Let observation $I$ be given by:
\[
I = \begin{array}{|c|c|c|}
\hline
0 & 1 & 0  \\ \hline
0 & {\pmb ?} & 1  \\ \hline
1 & 1 & {\pmb ?} \\ \hline
\end{array}.
\]
Then the set $\com(I)$ is given by:
\begin{multline*}
\com(I) = \left\{ 
\begin{array}{|c|c|c|}
\hline
0 & 1 & 0 \\ \hline
0 & {\pmb 0} & 1  \\ \hline
1 & 1 & {\pmb 0}  \\ \hline
\end{array},
\begin{array}{|c|c|c|}
\hline
0 & 1 & 0  \\ \hline
0 & {\pmb 0} & 1  \\ \hline
1 & 1 & {\pmb 1}  \\ \hline
\end{array},\right.\\
\left.
\begin{array}{|c|c|c|}
\hline
0 & 1 & 0  \\ \hline
0 & {\pmb 1} & 1  \\ \hline
1 & 1 & {\pmb 0}  \\ \hline
\end{array},
\begin{array}{|c|c|c|}
\hline
0 & 1 & 0  \\ \hline
0 & {\pmb 1} & 1  \\ \hline
1 & 1 & {\pmb 1} \\ \hline
\end{array}
\right\}.
\end{multline*}
As can be easily counted, $C(I) = 7$.\qed
\end{example}

We will say that a CA $A$ fits the observation $I$ if and only if there exists an $I'\in\com(I)$ and a sequence of natural numbers $(\tau_n)$ such that $\tau_n < \tau_{n+1}$ and for any $n\in\{1,2,\ldots,N-1\}$ it holds: 
\begin{equation}
A^{\tau_n}(I'[1]) = I'[n+1].
\end{equation}

\begin{prop}
\label{prop:alternative-prob}
Rule $A$ fits the observation $I$ if and only if there exist an $I'\in\com(I)$ and a sequence of natural numbers $(t_n)$ such that $t_n \leq t_{n+1}$ and for any $n\in\{1,2,\ldots,N-1\}$ it holds: 
\begin{equation}
A^{t_n}(I'[n]) = I'[n+1].
\end{equation}
\end{prop}

The sequence $(\tau_n)$ in the definition of fitting, corresponds to the time steps in the CA evolution (which are assigned to the rows of the observation), while the sequence $(t_n)$ in Proposition \ref{prop:alternative-prob} refers to the number of missing time frames between two consecutive rows in the observed diagram. Obviously $\tau_{n} = \sum_{i=1}^n t_i$.

In practice, it is useful to be able to use more than one observation for the identification. Therefore, we will consider observation sets $\mathcal{I}$ containing a finite number of observations. For simplicity, we assume that the elements of $\mathcal{I}$ are numbered, {\it i.e.}\ $\mathcal{I} =\{I_1,\dotsc,I_{\abs{\mathcal{I}}}\}$. 
We will say that rule $A$ fits the observation set $\mathcal{I}$, if it fits all of the observations in the set. 

Note that for the sake of simplicity we will write $C(\mathcal{I})$ to express the number of observed states in all of the observations belonging to $\mathcal{I}$, {\it i.e.} $C(\mathcal{I}) = \sum_{I\in\mathcal{I}} C(I)$. Additionally, we will write $M(\mathcal{I})$ to denote the total number of columns in all of the observations belonging to $\mathcal{I}$, {\it i.e.} $M(\mathcal{I}) = \sum_{I\in\mathcal{I}} M_I$ where $M_I$ is the number of columns of observation $I$.

For a non-empty observation set $\mathcal{I}$, the set $\rules(\mathcal{I})$ will denote all CA rules that fit the observation set $\mathcal{I}$. The identification problem is defined as finding the elements of the set $\rules(\mathcal{I})$ based on $\mathcal{I}$. In practice, our goal will be limited to finding at least one of the elements of $\rules(\mathcal{I}) \cap \mathcal{A}_r$ for some $r>0$. The problem can also be seen from the machine learning perspective in which the observation set is a training set, from which we try to learn and build a set of rules based on this knowledge.

The following fact will be used in the design of the identification algorithm, to simplify calculations. Informally, it could be expressed by understanding the observation set $\mathcal{I}$ as a set of conditions that the rule needs to meet. Having fewer conditions, it becomes more likely to find solutions meeting those conditions.

\begin{fact}
Let $\mathcal{I}$ be an observation set, and let $\mathcal{I}'\subset \mathcal{I}$. Then $\rules(\mathcal{I})\subset\rules(\mathcal{I}')$.\label{fac:rules-set}
\end{fact}

Since we consider only finite observation sets, we know that for every observation set $\mathcal{I}$ there exists a $T>0$ such that, if a solution exists, and $(t^I_n)$ is the time gap sequence of observation $I\in\mathcal{I}$, then $1\leq t^I_n\leq T$, for every $n$.
In the construction of the solution algorithm, we will assume that an upper-bound for $T$ is known.

\section{CA Identification as an optimization problem}
\label{sec:opt}
The identification problem, defined in Section \ref{sec:prob}, can be formulated as an optimization problem, which in turn enables the use of evolutionary search methods.

We start with an auxiliary definition. Let $a,b\in\{0,1,?\}^M$ be some vectors. We define the distance between $a$ and $b$ as:
\begin{equation}
\dist(a,b) = \sum_{a_i,b_i\in\{0,1\}}\abs{a_i-b_i}.
\end{equation}
We assume that if there is no $i$ such that $a_i \neq\ ?$ and $b_i \neq\ ?$ then $\dist(a,b)=0$. Therefore $\dist(a,b)=0 \not\Rightarrow a=b$.

Assume that $\mathcal{I}$ is a set of observations of some unknown CA belonging to $\mathcal{A}_r$, {\it i.e.}\ $\rules(\mathcal{I})\cup\mathcal{A}_r\neq\emptyset$. Let $A$ be a CA, and for every $I\in\mathcal{I}$, let $(\tau^I_n)$ be a strictly increasing sequence of natural numbers. 

As a start, we define the error measure $E_\mathcal{I}(A, (\tau^I_i))$, which measures how well a given CA $A$ fits the observation set $\mathcal{I}$, assuming that $\tau^I_i$ is the time step of the $i$--th row in observation $I$. The measure $E_\mathcal{I}$ is defined as:
\begin{equation}
E_\mathcal{I}(A, (\tau^I_i)) = \sum_{I\in\mathcal{I}} \sum_{n=1}^{N_I-1} \dist(A^{\tau^I_n}(I[1]), I[n+1]),
\label{eq:error0}
\end{equation}
where $N_I$ is the number of rows of observation $I\in\mathcal{I}$. The following fact is an direct consequence of the definition of the identification problem.
\begin{fact}
$A\in\rules(\mathcal{I})$ if and only if there exists a sequence $(\tau^I_i)$ such that $E_{\mathcal{I}}(A,(\tau^I_i)) = 0$.
\end{fact}

Note that in the case when $\mathcal{I} = \{I\}$ we will write $E_I$ instead of $E_{\{I\}}$.

 
Let $(t_i)$ be a sequence of natural numbers, and let $A$ be a CA rule. Observation $\bar{I}^A_{(t_i)}$  defined as:
\[
\bar{I}^A_{(t_i)}[n,m] = \begin{cases}
I[n,m], & \textrm{if}\ I[n,m]\neq\,?,\\
A^{t_{n-1}}(\bar{I}^A_{(t_i)}[n-1])[m], & \textrm{if}\ I[n,m]=\,?,
\end{cases}
\]
will be referred to as the $A$--completion of $I$ with time gaps $(t_i)$. Note that any observation $I$ satisfies $I[1] = \bar{I}^A_{(t_i)}[1]$ for any $A$, $(t_i)$. 

\begin{fact}
$\bar{I}^A_{(t_i)}\in\com(I)$.
\end{fact}

\begin{example}
\label{ex:ii}
Assume that CA $A$ is ECA 150 with LUT given by Table \ref{tab:lut-150} and local rule $f_{150}$. Let $(t_i)_{i=1}^2 = (1,2)$. We consider the observation $I$ defined in Example \ref{ex:i} and compute $\bar{I}^A_{(t_i)}$.
\[
I = \begin{array}{|c|c|c|}
\hline
0 & 1 & 0  \\ \hline
0 & {\pmb ?} & 1  \\ \hline
1 & 1 & {\pmb ?} \\ \hline
\end{array}\quad\quad
\bar{I}^A_{(t_i)} = \begin{array}{|c|c|c|}
\hline
0 & 1 & 0  \\ \hline
0 & {\pmb 1} & 1  \\ \hline
1 & 1 & {\pmb 0} \\ \hline
\end{array}\
\]
The calculation is as follows. Firstly we compute $\bar{I}^A_{(t_i)}[2,2]$. Since $t_1 = 1$ we simply apply the rule to the first row of $I$, {\it i.e.} $\bar{I}^A_{(t_i)}[2,2] = f_{150}(I[1,1], I[1,2], I[1,3]) = 1$. Since $t_2 = 2$, to find $\bar{I}^A_{(t_i)}[3,3]$, we first need to compute one additional configuration by evaluating the rule on configuration $\bar{I}^A_{(t_i)}[2]$. It is easy to check that $A(\bar{I}^A_{(t_i)}[2]) = (0,0,0)$, and thus $\bar{I}^A_{(t_i)}[3,3] = f_{150}(0,0,0) = 0.$\qed
\end{example}

Based on Proposition \ref{prop:alternative-prob}, we define an alternative error measure $\widetilde{E}_{\mathcal{I}}(A,(t^I_i))$ that will turn out to be more useful in the construction of the solution algorithm. Assuming that $(t^I_i)$ is a sequence of natural numbers representing time gaps, $\widetilde{E}_{\mathcal{I}}$ is defined as:
\begin{equation}
\widetilde{E}_\mathcal{I}(A, (t^I_i)) = \sum_{I\in\mathcal{I}} \sum_{n=1}^{N_I-1} \dist\Big(A^{t^I_n}\big(\bar{I}^A_{(t^I_i)}[n]\big), \bar{I}^A_{(t^I_i)}[n+1]\Big).
\end{equation}
Since $\bar{I}^A_{(t_i)}\in\com(I)$, we can express $\widetilde{E}_{\mathcal{I}}$ without using the function $\dist$ as:
\begin{equation}
\widetilde{E}_\mathcal{I}(A, (t^I_i)) = \sum_{I\in\mathcal{I}} \sum_{n=1}^{N_I-1} \abs{A^{t^I_n}\big(\bar{I}^A_{(t^I_i)}[n]\big) - \bar{I}^A_{(t^I_i)}[n+1]}.
\end{equation}

\begin{example}
We refer again to observation $I$, CA $A$ and $(t_i)$ used in Example \ref{ex:ii} and we compute the error measures $E_I$ and $\widetilde{E}_I$. Let us start with $E_I$. Following the fact that $\tau_n = \sum_{i=1}^n t_i$, we get $(\tau_i) = (1,3)$. The error measure $E_I$ can be computed easily by evolving $A$, starting from the initial configuration $I[1]$ and comparing the results with the values in $I$, for entries not occupied by $?$. 

Starting from the top: $A(I[1]) = (1,1,1)$. Since $\tau_1 = 1$ we compare the outcome with the second row of $I$. As we see, $I[2,1] = 0 \neq 1$ has an incorrect value, $I[2,2]=?$ so it does not contribute to the error and $I[2,3]=1$ which is a correct value. Since $\tau_2=3$ we should further evolve $A$ three times, starting from $A(I[1])$, but since $A((1,1,1)) = (1,1,1)$, we can simply compare $I[3]$ with $(1,1,1)$ and see that no errors occur. Summing up, the total error is: $E_I(A,(\tau_i)) = 1$. 

Similarly, we find the value of $\widetilde{E}_I$, by taking pairs of rows $\bar{I}^A_{(t_i)}[n]$ and $I[n+1]$ and comparing the results of $A^{t_n}(\bar{I}^A_{(t_i)}[n])$ and $I[n+1]$. The error in the first pair of rows is the same as in the case of $E_I$. For the second pair the initial condition is $\bar{I}^A_{(t_i)}[2] = (0,1,1)$, and since $A(0,1,1) = (0,0,0)$ and since $A((0,0,0)) = (0,0,0)$, we do not further evaluate $A$. We compare $(0,0,0)$ with $I[3]$, which yields 2 incorrect values. Summing up, the total error is $\widetilde{E}_I(A,(t_i)) = 3$. \qed
\end{example}

The relation between $E_\mathcal{I}$ and $\widetilde{E}_\mathcal{I}$ is expressed by the following proposition.

\begin{prop}
Let $A$ be a CA rule and $\mathcal{I}$ an observation set. There exists a strictly increasing sequence $(\tau^I_i)$ of natural numbers, such that $E_\mathcal{I}(A,(\tau^I_i))=0$ if and only if there exists a sequence $(t^I_i)$ of natural numbers such that $\widetilde{E}_\mathcal{I}(A,(t^I_i))=0$.\label{fac:relationoferror}
\end{prop}

As a consequence of Proposition \ref{fac:relationoferror}, the identification problem can be expressed mathematically as the minimization of $\widetilde{E}$. Note that this is only possible due to the assumption that observation set $\mathcal{I}$ contains partial space-time diagrams of some unknown CA. In a more general setting, where the observations could have a more complex origin, such a simplification is not possible.

As mentioned earlier, we consider the case where the upper bound for the time gaps is known. Using this knowledge, we define the error measure $\widetilde{E}_{\mathcal{I}}$ independently of the selection of $(t^I_i)$ as:
\begin{equation}
\widetilde{E}_\mathcal{I}(A) = \min_{\substack{(t^I_i)\\1\leq t^I_i \leq T}} \widetilde{E}_\mathcal{I}(A, (t^I_i)).
\label{eq:error}
\end{equation}

Note that the minimum in (\ref{eq:error}) is always defined, since there is a finite number of possibilities for the choice of $t^I_n$. Additionally, note that for a spatially complete observation $I$, the choice of $t^I_n$ is independent of the choice of $t^I_m$ for any $n\neq m$, and for observations $I$ and $J$, the choice of $(t^I_i)$ is independent from the choice of $(t^J_i)$. Consequently, to find the value of $\widetilde{E}_\mathcal{I}$ in the case of a spatially complete observation set, we need to examine at most $\sum_{I\in\mathcal{I}}T\,(N_I-1)$ sequences of time gap lengths.

In the general case, the choices of the values of $(t^I_i)$ are dependent on each other, and thus in order to find the exact value of the error measure we need to examine all of the $T^{N_I-1}$ possibilities, which holds a substantial computational burden. Due to this, even in the case of partial observations, we follow the approach described above and treat the time steps independently. The only difference that we introduce is that if for given $n$, few different candidate values for $t^I_n$ lead to the same, minimal value of the pairwise error, one of those candidates is being selected randomly. Such an approach, is a stochastic overestimation of the error, {\it i.e.} the calculated value will never be lower than the actual error. Additionally, if a given CA is a solution to the problem, recalculating the approximate error measure multiple times increases the probability of finding the exact value, which is found by taking the minimum of all of the obtained results. Such an approach turned out to be sufficient in the discussed context. 

\section{Evolutionary algorithm}
\label{sec:ga}
Having stated the identification problem as an optimization problem in this section, we describe its solution using an evolutionary algorithm based on the classical GA. In order to follow the GA approach, we need to define the individuals' representation, the population structure, a fitness function for ranking the individuals, but also the selection procedure for reproduction, and finally the cross-over and mutation operators. Formally, also halting conditions need to be formulated.

\subsection{Representation of individuals and population structure}
Here, the individuals that make up the population are CAs, encoded through the LUT of their local rules, which is possible since the LUT of any CA $A\in\mathcal{A}_r$ can be represented as a bit-string of length $2^{2\,r+1}$. We assume that the population consists of CA belonging to $\mathcal{A}_r$, for some $r>0$.

We consider populations of $P>0$ individuals. By $\mathcal{P}^i$ we denote the population of the $i$--th generation of the GA. The population $\mathcal{P}^1$ is the initial population, and is constructed by randomly selecting $P$ bit-strings. Populations $\mathcal{P}^i$ for $i>1$ are the outcomes of applying the genetic operators, according to the rules described in the remainder of this section.

\subsection{Fitness function}
\label{sec:fitness}
The fitness function is directly related to the error measure $\widetilde{E}_{\mathcal{I}}$ defined by (\ref{eq:error}). Although Proposition \ref{fac:relationoferror} states that the error measures given by (\ref{eq:error0}) and (\ref{eq:error}) can be used interchangeably, preliminary experiments showed that the later results in efficient and convergent algorithm, while suboptimal results were obtained using the measure given by (\ref{eq:error0}). This follows from the fact that the error in row $n$ is affected by errors appearing in rows $2,\dotsc,n-1$. As we know from the research on dynamical properties of CAs, small initial perturbations can strongly affect the final system state \cite{jan.lyap}. For that reason, it is easier to optimize $\widetilde{E}_{\mathcal{I}}$ with a GA as compared to $E_{\mathcal{I}}$. 

Let $L\in\{0,1\}^{2^{2\,r+1}}$ be a LUT of some local rule which defines a CA $A$. Then $\fit_{\mathcal{I}}(L)$ denotes the fitness of $A$, and is defined as:
\begin{equation}
\fit_{\mathcal{I}}(L) = C(\mathcal{I}) - M(\mathcal{I}) - \widetilde{E}_\mathcal{I}(A) .
\label{eq:fit-org}
\end{equation}
The fitness function takes integer values from 0 up to $C(\mathcal{I}) - M(\mathcal{I})$, {\it i.e.} there are finitely many possible values of the fitness function. The goal of the GA is to maximize fitness, and a CA with a maximal fitness value is a solution of the identification problem. From the above, it is clear that if $C(\mathcal{I}) - M(\mathcal{I})$ is close to zero, solving the problem is infeasible, since the number of possible values is very small and the population is not able to gradually increase its fitness. Additionally, if $C(\mathcal{I}) = M(\mathcal{I})$, then the problem is trivial because every CA is a solution.


The fitness defined by (\ref{eq:fit-org}) has proven to work effectively, but the computing time needed for its evolution becomes unacceptable if the observation set is large. Therefore, during the evolution, to estimate the value of $\fit_{\mathcal{I}}$ we use $\fit_{\mathcal{I}'}$ for some non-empty subset $\mathcal{I}'\subset\mathcal{I}$. We start by randomly selecting elements for the subset $\mathcal{I}'$. Subsequently, but before evolving a new population we replace one of the elements in the subset $\mathcal{I}'$ with a randomly selected observation from $\mathcal{I}$. Due to Fact \ref{fac:rules-set} we are sure that such an approach does not result in reducing the solution set. 

\subsection{Selection operator}
\label{sec:select}
Having defined the fitness function, we can define the selection operator, which is responsible for selecting the parent individuals that will be used to produce the next generation. We use a random selection method where the selection probability of a given individual is proportional to its fitness. Individuals are selected with replacement, {\it i.e.}\ individuals might be selected multiple times for reproduction.
\subsection{Cross-over operator}
To produce offspring, we select two parents according to the procedure described in Subsection \ref{sec:select}. A uniform cross-over operator is used, {\it i.e.} if $L_1, L_2$ denote parents, the outcome of the cross-over operator is a vector $L_c$ with values that are randomly selected from $L_1$ and $L_2$, {\it i.e.}\ $\mathbb{P}(L_c[i] = L_1[i]) = \mathbb{P}(L_c[i] = L_2[i]) = 0.5$.


\subsection{Mutation operator}
Finally, the offspring individual is mutated. A simple bit-flip mutation is being used, {\it i.e.} for every position of the vector a decision is made whether or not the value should be flipped, with $p_f$ being the probability of flipping the value. The expected number of flipped positions in the population is $P\,p_f\,2^{2\,r+1}$.

\subsection{Elite survival}
After evolving a new population, the elite survival procedure is applied. Our experiments proved that such an approach is required to reach convergence. The procedure is implemented by a deterministic selection of $P_E \ll P$ fittest individuals from the previous population used to replace randomly selected individuals in the newly evolved one.

Including this elite survival process can dramatically increase the performance of the algorithm, though there are cases where such an approach causes the population to progress towards a local optimum. To overcome this, we apply a simple, adaptive procedure that deactivates elite survival in cases when the maximum fitness value of the population remained constant for more than $N_{\textrm{off}}$ generations. The elite survival procedure is again switched on after a predefined number of $N_{\textrm{on}}$ generations, or if the maximum fitness improved.

\subsection{Halting conditions}
\label{sec:halt}
The algorithm evolves by generating populations according to the procedure described above until a maximum, predefined number of populations $\Lambda$ was evolved or, if a CA that fits the observation set was discovered. 

As mentioned in Subsection \ref{sec:fitness} during the evolution, the fitness $\fit_{\mathcal{I}}$ is approximated by $\fit_{\mathcal{I}'}$ for some $\mathcal{I}'\subsetneq\mathcal{I}$, which is effective for selection, but can not be used in the halting condition since $\fit_{\mathcal{I}'}(A) = C(\mathcal{I}') - M(\mathcal{I}')$ does not imply $\fit_{\mathcal{I}}(A) = C(\mathcal{I}) - M(\mathcal{I})$. Therefore, for the individual $A$ with the highest value $\fit_{\mathcal{I}'}(A)$, we additionally calculate $\fit_{\mathcal{I}}(A)$ and base the halting condition on it, {\it i.e.} the algorithm stops as soon an element is found. 

\section{Results of experiments}
\label{sec:res}
By means of our experiments we verified to what extent the partiality of observations affects the efficiency of the GA in terms of the number of GA iterations required to find a solution. 

We concentrated on two ECAs: 150 and 180, with LUTs given in Table \ref{tab:lut-150} and \ref{tab:lut-180}, respectively. 

\begin{table}[ht]
\renewcommand{\arraystretch}{1.3}
\caption{LUT of ECA 180}
\label{tab:lut-180}
\centering
\begin{tabular}{|>{$}c<{$}|>{$}c<{$}|>{$}c<{$}|>{$}c<{$}|>{$}c<{$}|>{$}c<{$}|>{$}c<{$}|>{$}c<{$}|}
\hline
111 & 110 & 101 & 100 & 011 & 010 & 001 & 000 \\
\hline
1 & 0 & 1 & 0 & 1 & 0 & 1 & 0 \\
\hline
\end{tabular}
\end{table}

In this experiment, the GA evolution is based on observation sets $\mathcal{I}_A(k)$ for $k=\{0,1,\dotsc,150\}$ and ECA $A\in\{150, 180\}$. The integer $k$ will be referred to as the problem number. The observation set $\mathcal{I}_A(0)$ is a set of $\Omega>0$ observations obtained from $\Omega$ different, random initial conditions common for both $A$, by selecting subsequent configurations of ECA $A$ generating time gaps of random length from 1 to $T$. The set $\mathcal{I}_A(k)$ for $k>0$ is built from observations belonging to $\mathcal{I}_A(k-1)$ by modifying them in such a way that $\pi=2000$ randomly selected, completely observed entries are replaced by ``?''. In other words, by increasing $k$ the effect of spatial partiality is increased. As a result of such a procedure we obtained a series of observation sets $\big(I_A(k)\big)_{k=0}^{150}$, for which it holds $C(I_A(k)) - C(I_A(k+1)) = \pi$. The identification algorithm was then executed for each of the obtained observation sets.

Given that the family of ECAs contains only 256 members, the identification problem would be relatively easy to tackle, so we set the radius $r=2$, {\it i.e.} the population contains local rules with radius $r=2$ represented as bit-strings of length 32. Without this modification the algorithm is able to find a solution in a few iterations, by examining the entire search space.

In order to account for the stochastic nature of the GA, the experiment is repeated $L>0$ times for each $r$, $k$. The values of of the GA parameters used in our experiment setup are shown in Table~\ref{tab:exp1-params}.

\begin{table}[ht]
\caption{Parameters used in the experiment}
\label{tab:exp1-params}
\centering
\begin{tabular}{>{$}c<{$}|>{$}c<{$}|l}
\textrm{{\bf param}} & \textrm{{\bf value}} & {\bf description} \\ \hline
r & 2 & rule radius \\\hline
p_f & 0.01 & probability of flipping 1-bit in mutation \\
P & 512 & number of individuals in population \\
P_E & 32 & elite size \\ \hline
T & 10 & bound for the time gap length \\
C & 69 & number of rows / columns in each observation \\ 
\Omega & 64 & number of observations \\
\pi & 2000 & number of cells being removed from each observation set \\
s & 8 & number of samples for fitness approximation \\ \hline
\Lambda & 5000 & maximal number of the GA populations \\
L & 20 & number of repetitions of the GA per rule \\
\end{tabular}
\end{table}

\begin{figure}[!t]
\centering
\includegraphics[width=2.5in]{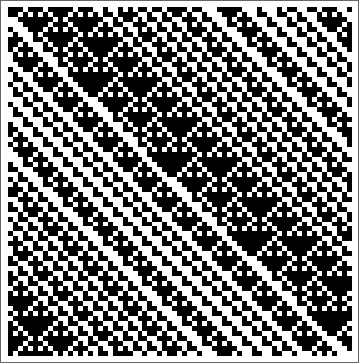}
\caption{Space-time diagram of ECA 180}
\label{fig:spatio2}
\end{figure}

The results vary significantly depending on the rule in question, which is not surprising since the dynamics of ECAs 150 and 180 is different. The normalized Maximum Lyapunov Exponent (nMLE) \cite{Wolfram:84,Sher:92,citeulike:9312129} of the former is the highest among all of the ECAs, and thereby this CA's behavior may be considered complex. In contrast, the nMLE of ECA 180 is only approximately 0.48, which hints that, in some sense, the behavior of this ECA is simpler than the one displayed by ECA 150. The differences in the overall dynamical complexity of these two CAs can be acknowledged by examining their space-time diagrams, which are depicted in Fig.\ \ref{fig:spatio1} and Fig.\ \ref{fig:spatio2}.

\begin{figure}[!t]
\centering
\subfloat[ECA 150]{\includegraphics[width=3.5in]{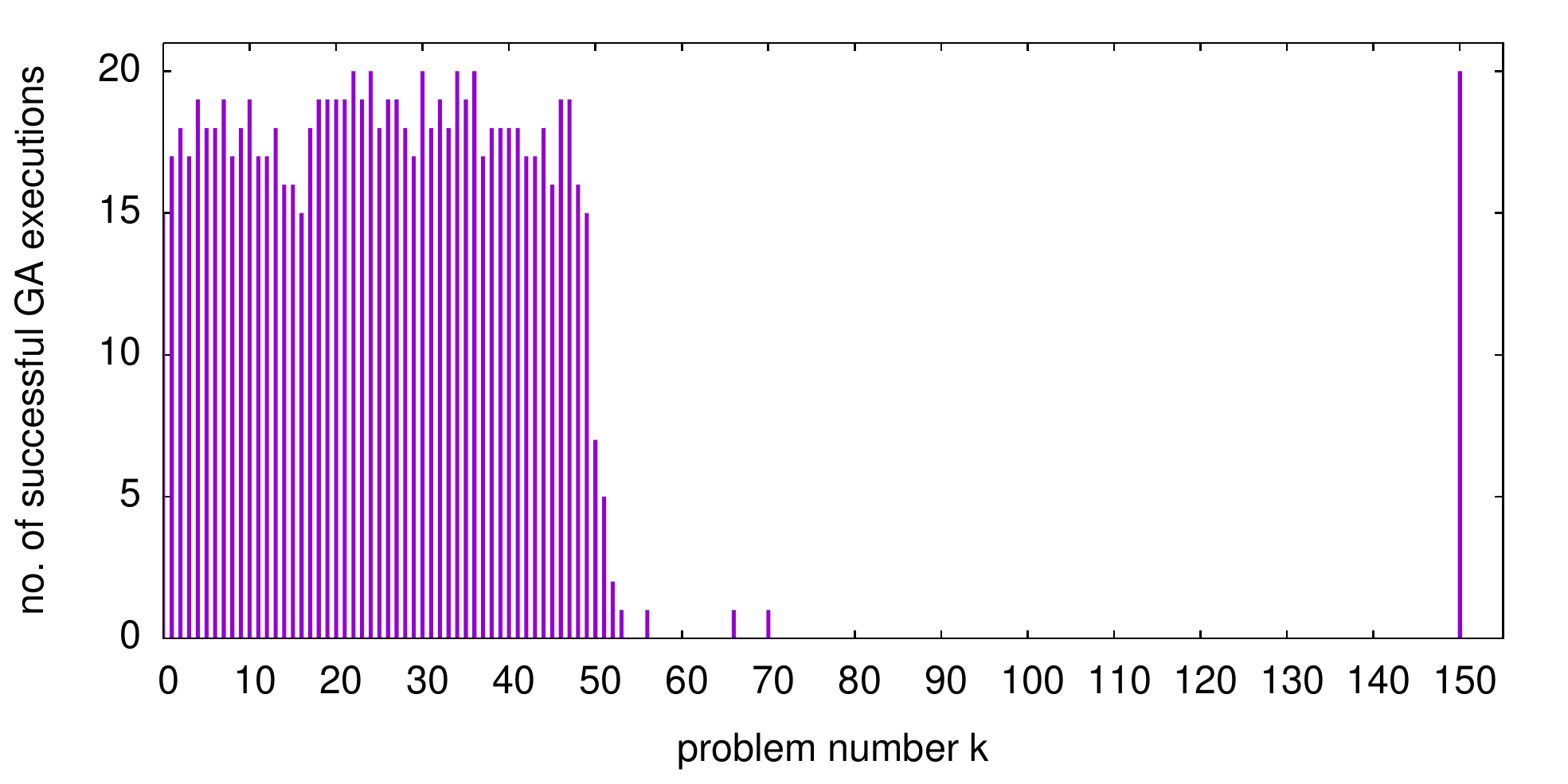}}\\
\subfloat[ECA 180]{\includegraphics[width=3.5in]{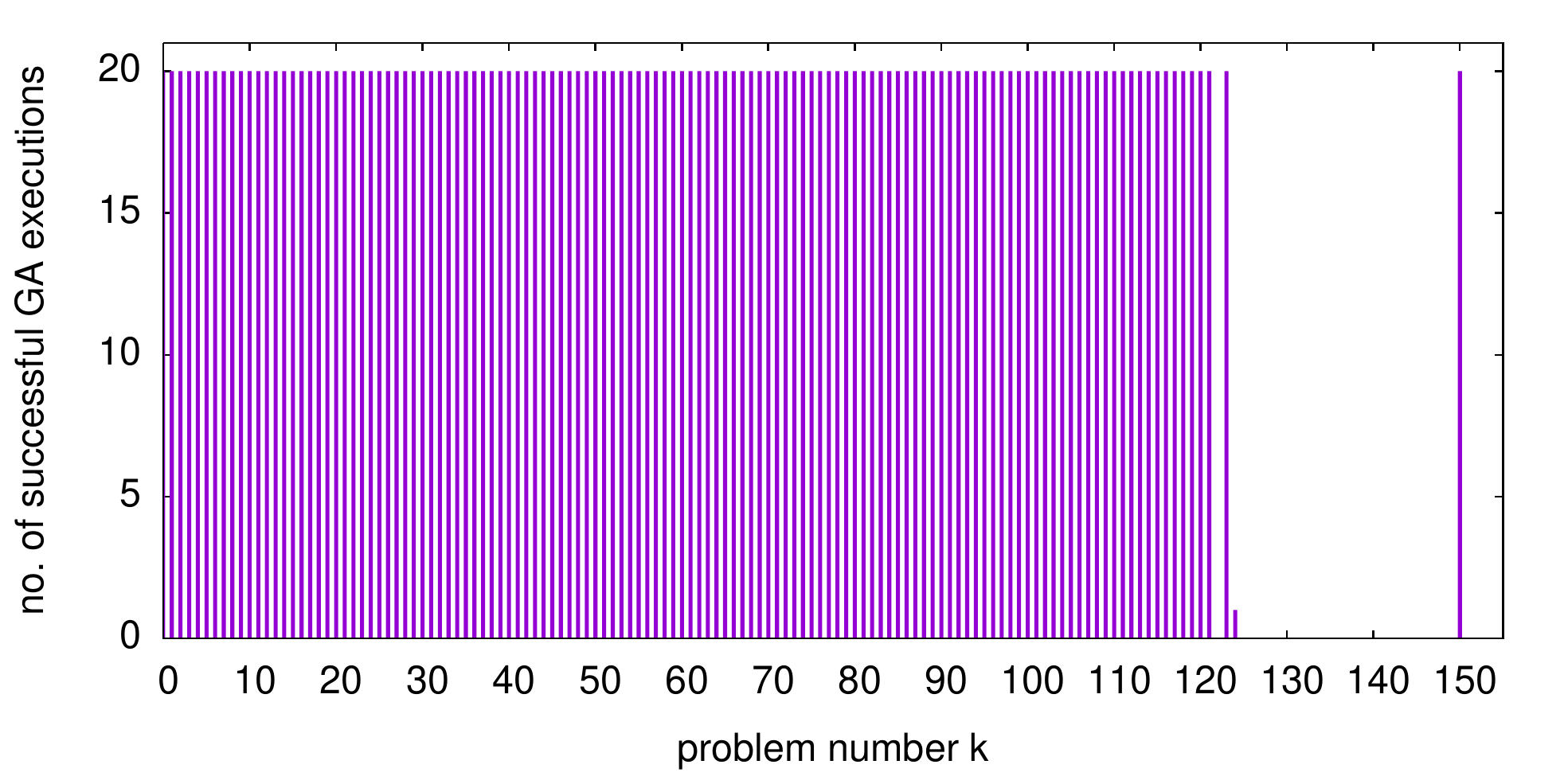}}
\caption{Number of successful GA executions}
\label{fig:res180-no}
\end{figure}

To understand the performance of the GA, we first checked for which $k$ the algorithm was able to find a solution (Fig.\ \ref{fig:res180-no}). When comparing the plot for ECA 150 with the one for ECA 180, it is clear that the identification problem turned out to be much more challenging for ECA 150. Indeed, for this ECA, the algorithm was effective only if the number removed observation elements was smaller than $50\,\pi$, whereas it mostly failed when more spatial partiality was added. Besides, even for $k$ close to 0, not all of the GA executions were successful. In contrast, identifying ECA 180 was always possible for $k<120$, but for $k>120$ we see a sudden drop in the performance. Note that in both cases, for $k=150$ a solution was easily found, since for this setting the problem is trivial, {\it i.e.} almost all CAs can be considered a solution.

The above results suggest that, depending on the dynamical characteristics of the CA in question, the maximum allowable number of missing elements in the observations differs. Further research is undertaken to better understand the link between the identifiability and dynamics of CAs.

\begin{figure}[!t]
\centering
\subfloat[ECA 150]{\includegraphics[width=3.5in]{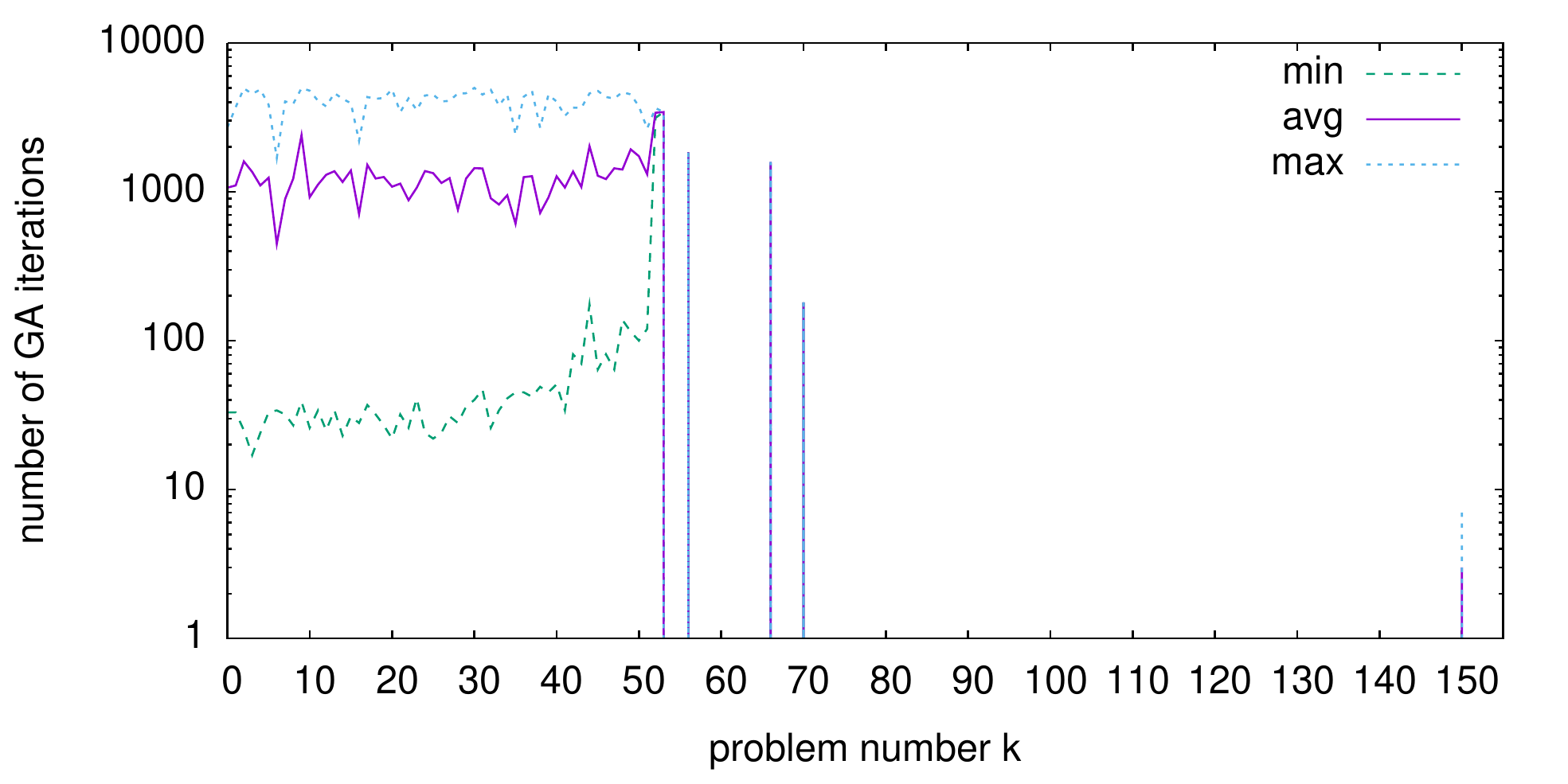}}\\
\subfloat[ECA 180]{\includegraphics[width=3.5in]{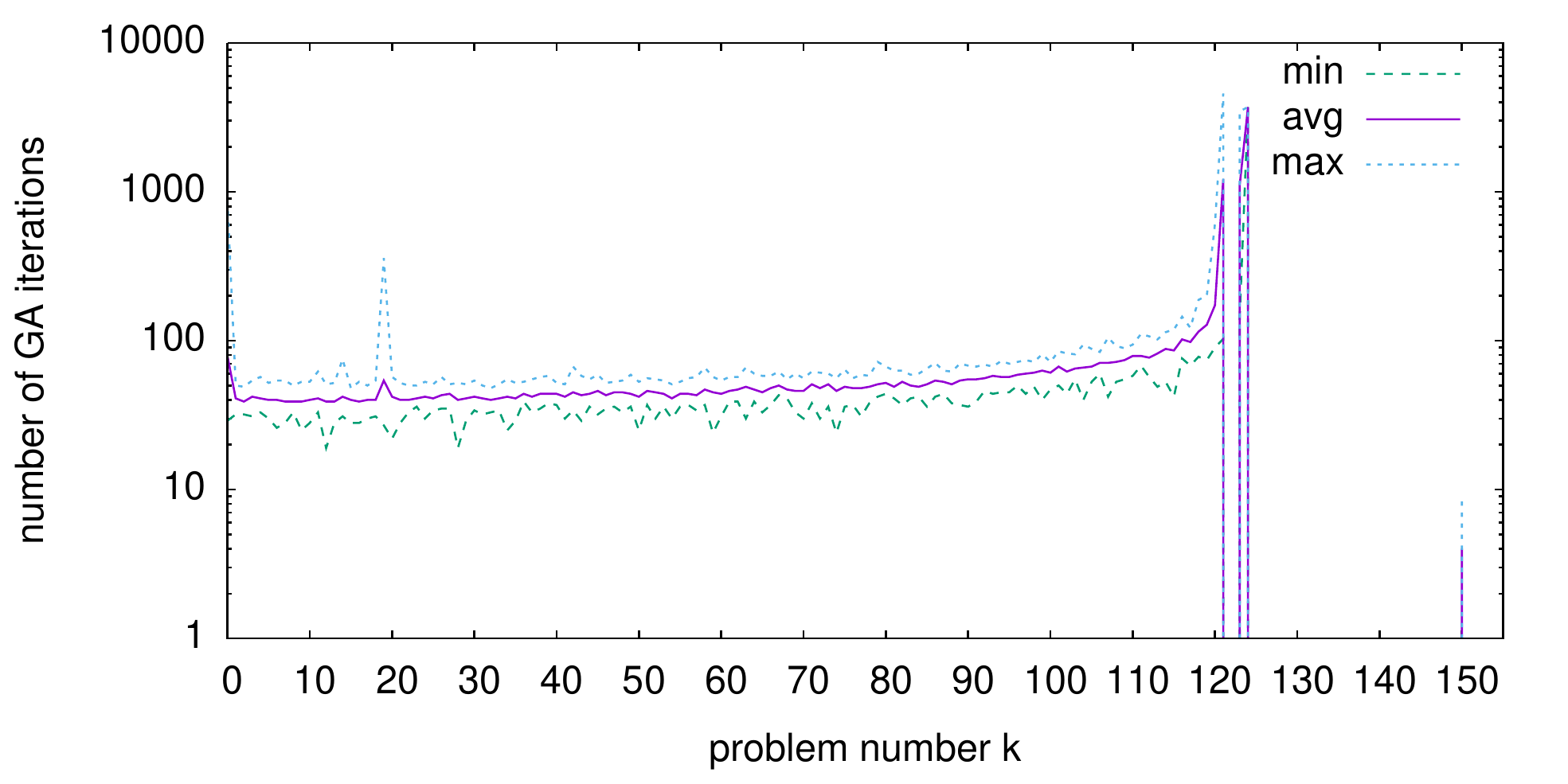}}
\caption{Number of GA iterations required to find a solution}
\label{fig:res180}
\end{figure}

Figure \ref{fig:res180} depict the minimum, average and maximum number of GA iterations among the runs resulting in a solution for ECA 150 and ECA 180, respectively. In the case of ECA 180, we see that the efforts needed for finding a solution grows as $k$ increases, up to the point where it becomes impossible. Furthermore, we see that in most cases the difference between maximal and minimal values is relatively low. In the case of ECA 150, the results are much less stable. The differences between maximal and minimal values are substantial, and the efforts needed to find the solution do not steadily grows with the growing spatial partiality. The only similarity between the two CAs seems to be in the fact that there exists some critical $k$ beyond which the problem becomes impossible to solve. 

\section*{Summary}
In this paper we introduced the identification problem of CAs in the context of partial observations. An evolutionary algorithm for tackling the problem was presented, and its performance was verified for the two ECAs. The initial experiments suggest that the difficulty of the identification problem is somehow linked to the dynamical complexity of the CAs. The problem and solution algorithm presented in this paper, should be considered as one of the first steps in identifying CAs from data originating from real-world phenomenon observations. Unavoidably, such observations will be somehow incomplete in the sense that it is impossible to continuously track the involved processes. 

\IEEEtriggeratref{8}
\bibliographystyle{IEEEtran}
\bibliography{ieee/IEEEabrv,identify}
\end{document}